\documentclass[fleqn,10pt]{wlscirep}
\usepackage[utf8]{inputenc}
\usepackage[T1]{fontenc}
\usepackage{floatrow}
\usepackage{makecell}
\usepackage{caption}
\usepackage{subcaption}

\title{Medical Scientific Table-to-Text Generation with Human-in-the-Loop under the Data Sparsity Constraint}

\author[1,=]{Heng-Yi Wu}
\author[2,=]{Jingqing Zhang}
\author[2]{Julia Ive}
\author[2]{Tong Li}
\author[2]{Vibhor Gupta}
\author[1,*]{Bingyuan Chen}
\author[2,**]{Yike Guo}
\affil[1]{Development Science Informatics, Genentech Inc, USA}
\affil[2]{Pangaea Data Limited, UK, USA}
\affil[=]{These authors contributed equally to this work.}
\affil[*]{chenb14@gene.com}
\affil[**]{yguo@pangaeadata.ai}

\begin{abstract}
Structured (tabular) data in the preclinical and clinical domains contains valuable information about individuals and an efficient table-to-text summarization system can drastically reduce manual efforts to condense this data into reports. However, in practice, the problem is heavily impeded by the data paucity, data sparsity and inability of the state-of-the-art natural language generation models (including T5, PEGASUS and GPT-Neo) to produce accurate and reliable outputs. In this paper, we propose a novel table-to-text approach and tackle these problems with a novel two-step architecture which is enhanced by auto-correction, copy mechanism and synthetic data augmentation. The study shows that the proposed approach selects salient biomedical entities and values from structured data with improved precision (up to 0.13 absolute increase) of copying the tabular values to generate coherent and accurate text for assay validation reports and toxicology reports. Moreover, we also demonstrate a light-weight adaptation of the proposed system to new datasets by fine-tuning with as little as 40\% training examples. The outputs of our model are validated by human experts in the Human-in-the-Loop scenario.
\end{abstract}
\begin{document}

\flushbottom
\maketitle
 
\thispagestyle{empty}

\section*{Introduction}

General data-to-text (Data2Text) generation with neural methods is a widely studied problem with a range of successful approaches \cite{chang-etal-2021-neural,chang-etal-2021-order,puduppully2021datatotext}. Table-to-text (Table2Text) generation is its sub-problem focused on extracting relevant tabular values, describing (extractive reasoning) and deducing conclusions about them (abstractive reasoning) in the text \cite{wang-etal-2020-towards,gong-etal-2020-tablegpt,chen-etal-2020-shot}. Despite the majority of the approaches being quite successful, there are still several major challenges to embrace to increase the chances of successful deployment of these models in real-life scenarios. State-of-the-art (SOTA) text generation models in general and Table2Text models, in particular, are Transformer-based \cite{vasvani:nips:2018} and require significantly more data for proper training than traditional approaches. In practice there is a scarcity of such at the volumes which will be required; hence models struggle to learn how to copy data from inputs to outputs correctly. Higher precision is usually achieved by adapting these models to specific tasks via fine-tuning, which creates practical issues of maintaining these big models when many different tasks are considered. Moreover, the models operate as black boxes with little ability to convey reasons for their decisions.  

Automated Table2Text generation in the medical scientific domain has the potential to drastically reduce the burden of human experts spending a lot of their time writing reports. However, the aforementioned challenges are particularly relevant for the medical scientific domain. It is essential for the validity of the generated text that the precision of the tabular values is preserved to a relevant degree. Model decisions need to be transparent for clinical experts to rely on them. In addition, data sparsity is a major bottleneck in the medical scientific domain: texts are of high quality, however they are produced in small quantities by experts. Medical reports tend to vary significantly with respect to the type of information they summarise and the aspects of the input they reflect. In such conditions, it is almost infeasible to create a model per text type. Hence neural Table2Text generation of the scientific medical text remains an understudied problem \cite{Pauws2019}. 

In this work, we pioneer the task of Table2Text generation in the medical scientific domain and address the aforementioned challenges with a two-step generation approach that enhances the effect of augmented data with copy mechanisms \cite{see-etal-2017-get,vinyals2015pointer} to increase precision and interpretability of the resulting Table2Text model. 

Following best practices \cite{puduppully2021datatotext}, our framework consists of two components: the \textbf{Table Extractor} which identifies tabular values that are to be described in the text and the \textbf{Text Generator} that takes the extract (as produced at the previous step) and generates fluent text. Both components are SOTA language models \cite{zhang2019PEGASUS} equipped with copy mechanisms that map the values directly from the inputs (acting as pointers). As the Table2Text task usually assumes the presence of values from the input structured data in the textual outputs, the straightforward data augmentation approach consists in building the training examples by slot-value replacements respecting the slot type \cite{chang-etal-2021-neural}: e.g., ``100 ml'' will be replaced by similar values occurring in similar conditions ``200 ml'' or ``300 ml''. We extend this approach further and increase the variety of numeral values by slight randomisation (by taking the value very close to the original value). Coupled with our copy mechanisms, our data augmentation procedure promotes training examples where only the matched values are changing and the copy mechanism allows the model to focus on them and ensure the precision of their copy. 

Furthermore, we introduce a Human-in-the-Loop protocol exploiting the best practices of Active Learning \cite{activelearning} and on-the-fly model adaptation. Active Learning relates a group of methods for choosing the training samples that are most likely to improve a system. The initial goal of these techniques is to save human effort by reducing the number of instances for human to validate. In this setting, we only select for validation those examples that will be most beneficial for model adaptation according to the model uncertainty in its output and thus ensure the minimisation of human corrective effort. Furthermore, by only fine-tuning an Automatic Corrector (\textbf{Autocorrector}) model built on the top of the main model we solve the practical issue of the costly large model adaptation. Our choice of Autocorrector architecture is informed by best practise in the field of Machine Translation domain~\cite{chatterjee-EtAl:2020:WMT}. Such Autocorrectors are able to perform direct changes to outputs and are easy to adapt \cite{parton-etal-2012-automatic}. 

Our main {\bf contributions} are: (i) proposal for a transparent Table2Text architecture for the generation of accurate medical scientific text with traceability to identify the link between automated narratives and values;
(ii) generalisable methodology of synthetic data generation to address the crucial issue of data sparsity for the medical scientific Table2Text;
(iii) by leveraging automatic correction techniques we reduce the cost of adaptation to new types of data by minimising human corrective intervention in the Human-in-the-Loop protocol.

We benchmark our approach on two types of Table2Text medical scientific data: assay validation reports and toxicology reports. Our approach outperforms the state-of-the-art methods by up to 0.13 absolute increase on Table Recall of copying tabular values accurately to text. Besides, we also provide an in-depth analysis of our model outputs as a result of the Human-in-the-Loop assessment and show the proposed approach adapts to new datasets with as little as 40\% training examples. 
\section*{Results}

\subsection*{Data}

\subsubsection*{Assay Validation Reports}

Assay Validation Report (AVR) is the comprehensive experiment report that evaluates and documents the quantitative performance of an assay, including sensitivity, specificity, accuracy, precision, detection limit, range and limits of quantitation. Full assay validation will include inter-assay and inter-laboratory assessment of assay repeatability and robustness. 

In this project, we extract 1,239 table-paragraph pairs from 92 raw assay validation reports. 106 pairs are reserved for testing and the other 1133 pairs are used for training as shown in Table~\ref{table:data-stat-bas}. The paragraphs describing tables are automatically paired from longer PDF reports (which are transformed to text by using OCR) by first matching the table number and then selecting the paragraph that contains the most values from the table. For better evaluation, tuplets from the test set are manually updated after this automatic selection procedure, as most of the errors of the automatic selection procedure are related to incomplete paragraphs split across PDF pages.   

In practice, the Assay Validation Reports can be continuously collected from different data vendors. Therefore, to evaluate the capability of our model to adapt to new data sources efficiently, we collect extra 218 and 219 samples for fine-tuning and testing respectively in the low-resource and light-weight Human-in-the-Loop scenario. The table-paragraph pairs used in the Human-in-the-Loop (HIL) scenario are manually curated and validated by one in-domain expert.

\begin{table}[!h]
\begin{center}
\renewcommand{\arraystretch}{1.1}
\scalebox{0.9}{
\begin{tabular}{|c|c|c|c|c|c|c|}
\hline
& \makecell{AVR \\ (training)}     & \makecell{AVR \\ (testing)} &  \makecell{AVR for HIL \\ (fine-tuning)}  & \makecell{AVR for HIL \\ (testing)} & \makecell{Toxicology Reports \\ (training)}     & \makecell{Toxicology Reports \\ (testing)} \\ \hline
\#, input tables            & 1133       & 106        & 218       & 219   & 43 & 44   \\ \hline       
avg \#, input words          & 185.8     & 185.1     & 198.6     & 204.7  & 32.3     & 38.6  \\ \hline
avg \#, input tokens     & 307.0      & 308.6       & 321.0       & 342.6  & 62.5     & 73.7  \\ \hline
\#, target reports     & 1133      & 106       & 218       & 219  & 43      & 44  \\ \hline
avg \#, words in reports           & 94.6       & 125.2       & 94.7       & 100.2   & 20.7       & 25.8  \\ \hline
avg \#, tokens in reports          & 126.8     & 170.0        & 125.4       & 135.5  & 43.8     & 53.0  \\ \hline
\end{tabular}}
\end{center}
\caption{\label{table:data-stat-bas} Statistics of Assay Validation Reports (AVR) and Toxicology Reports. The Assay Validation Reports include raw data for regular training and testing, and curated data for fine-tuning and testing in the Human-in-the-Loop (HIL) scenario. The real Toxicology Reports are split into training and testing and the training pairs are augmented by 43,000 synthetic training pairs. Also, due to limited size of real data, the Toxicology Reports are not used in the Human-in-the-Loop experiments.}
\end{table}

\subsubsection*{Toxicology Reports}

Toxicology Reports describing the results of preclinical toxicology studies carried out by pharmaceutical companies have been identified as a valuable source of information on safety findings for investigational drugs. In this project, Toxicology Reports which are abbreviated by experts and describe the findings in bullet points are used for model training and evaluation.

We extract 87 table-paragraph pairs (including the findings for body weight changes, clinical observations and mortality rates) from raw Toxicology Reports. 43 pairs are used for training and 44 pairs are reserved for testing as shown in Table \ref{table:data-stat-bas}. All pairs are manually curated and validated by one in-domain expert. However, as the number of real training pairs is rather limited, we generate 43,000 synthetic training pairs based on the 43 real training pairs to strengthen the robustness of model training. Also, due to the limited size of real data, Toxicology Reports are not included in the Human-in-the-Loop experiments.

% \begin{table}[!h]
% \begin{center}
% \renewcommand{\arraystretch}{1.1}
% \scalebox{0.73}{
% \begin{tabular}{|c|c|c|}
% \hline
% & Toxicology Reports (training)     & Toxicology Reports (testing) \\ \hline
% \#, input tables & 29 & 30     \\ \hline       
% avg \#, input words  & 34.3     & 40.4     \\ \hline
% avg \#, input tokens     & 65.9     & 75.6    \\ \hline
% \#, target reports     & 29      & 30     \\ \hline
% avg \#, words in reports           & 22.4       & 28.1 \\ \hline
% avg \#, tokens in reports          & 46.5     & 55.3  \\ \hline
% \end{tabular}}
% \end{center}
% \caption{\label{table:data-stat-toxicology} Statistics over the datasets of Toxicology: Body Weight Change and Clinical Observation}
% \end{table}

% \subsection*{PTPK Reports}

% TBD whether to add

% \red{TO DO Genentech}

% \red{TO DO Tong stats}

\subsection*{Evaluation Metrics} 

Considering the importance of the copying precision of tabular values in the resulting text, we have used two metrics focused on the assessment of the precision of the values in text reports copied from input tables.

\textbf{Table Recall} estimates the percentage of unique table extract values that appear in the generated text. This metric does not take the order and count of values in the table extract into account.
 
\textbf{BLEU Extract} is based on the popular text generation metric BLEU~\cite{Papieni02bleu} and computes the precision values of consecutive token spans between the table extract restored from the generated report and the reference table extract. The table extracts restored from the text contains only the words and numbers present both in the generated text and in the reference table extract. This metric accounts for the order and count of the tabular values in the extracts.

We also report standard metrics comparing lexical content of target and generated text: ROUGE, BLEU and TER. ROUGE \cite{Lin:2004} measures the overlapped n-grams between generated and target text and BLEU \cite{Papieni02bleu} measures \emph{n}-gram precision. Whereas, TER \cite{snover-AMTA-2006} measures the minimum number of edits (substitution, insertion, deletion and shift of a word) required to change a generated sentence so that it exactly matches a genuine one, so it is the lower the better.

\subsection*{Comparison to the State-of-the-Art} 

\begin{table}[h]
\renewcommand{\arraystretch}{1.1}
\begin{subtable}[h]{\textwidth}
\centering
\scalebox{0.9}{
\begin{tabular}{|l|c|c|c|c|c|c|c|}
\hline
& Table Recall $\uparrow$ & BLEU Extract $\uparrow$ & ROUGE 1 $\uparrow$ & ROUGE 2 $\uparrow$ & ROUGE L $\uparrow$ & BLEU $\uparrow$ & TER $\downarrow$ \\ 
\hline
Template & 0.3856 & 27.07 & 0.5666 & 0.4192 & 0.4984 & 40.30 & 0.6661\\ \hline
Content Selection and Planning & 0.5352 & 20.38 & 0.3996 & 0.2318 & 0.3140 & 17.34 & 1.4980\\ \hline
GPT-Neo & \textbf{0.8577} & 23.82 & 0.2110 & 0.0715 & 0.1207 & 1.83 & 1.3508\\\hline
T5 & 0.4923 & 15.87 & 0.3968 & 0.1956 &0.2435 & 14.82 & 1.1228\\\hline
PEGASUS & 0.5927 & 39.49 & 0.6128 & 0.4927 & 0.5550 &42.04 & 0.7147 \\\hline
\hline
Ours & \textbf{0.7245} & \textbf{46.92} & \textbf{0.6590} &\textbf{0.5304} & \textbf{0.5986} & \textbf{44.36} & \textbf{0.6455}\\ \hline
\end{tabular}}
\caption{\label{table:avr_sota} Results on Assay Validation Reports.}
\end{subtable}
\hfill
\begin{subtable}[h]{\textwidth}
\centering
\scalebox{0.9}{
\begin{tabular}{|l|c|c|c|c|c|c|c|}
\hline
& Table Recall $\uparrow$ & BLEU Extract $\uparrow$ & ROUGE 1 $\uparrow$ & ROUGE 2 $\uparrow$ & ROUGE L $\uparrow$ & BLEU $\uparrow$ & TER $\downarrow$ \\ 
\hline
GPT-Neo &  0.9127 & 25.03 & 0.6980 & 0.2867 & 0.5132 & 8.72 & 1.2586 \\\hline
T5 & 0.8784 & 26.47 & 0.7364 & 0.5874 & 0.7240 & 41.79 & 0.5990 \\\hline
PEGASUS & \textbf{0.8999} & \textbf{29.98} & 0.9038 & \textbf{0.7799} & 0.8867 & 64.71 & 0.2789 \\\hline
\hline
Ours & 0.8897 & 29.49 & \textbf{0.9055} & 0.7760 & \textbf{0.9034} & \textbf{65.18} & \textbf{0.2640}\\\hline
\end{tabular}}
\caption{\label{table:tox_sota} Results on Toxicology Reports.}
\end{subtable}
\caption{Comparison of the proposed approach with the state-of-the-art baseline methods on the test set of Assay Validation Reports and Toxicology Reports. The proposed approach outperforms all baseline methods on Assay Validation Reports and generate overall comparable results on Toxicology Reports. As the real data of Toxicology Reports is rather limited, in order to generate reasonable outputs, the baseline methods are trained by synthetic training pairs which are created by our proposed method. We observe GPT-Neo exhibits an exaggerated Table Recall but very low ROUGE and BLEU on both datasets due to the fact that the model tends to simply copy input tables into the output.}
\label{tab:all_sota_results}
\end{table}

In this section, we compare the proposed approach to the state-of-the-art (SOTA) methods. Apart from the baselines for Toxicology Reports, all other model-based baselines are trained with only real data without using synthetic data if it is not specified otherwise. The following baseline methods are considered:
\begin{itemize}
    \item \textbf{Template-based algorithm} is designed to find the closest matches between real training set and real testing set. For each test table, we select the closest template from the training data according to table titles using the longest contiguous matching sub-sequence algorithm which is defined as follows:
    $$D_{ro} = \frac{2K_m}{\left| S_1 \right| + \left| S_2 \right|}$$
    where $\left| S_1 \right|$, $\left| S_2 \right|$ are the length of two strings and $K_m$ represents the number of matching characters between two strings.
    We then fill the slots in the selected template with values from the corresponding positions in the test table.
    \item \textbf{PEGASUS} is a Transformer encoder-decoder model with the pre-training objective that masks and generates important gap sentences from an input document and achieves state-of-the-art performance on 12 abstractive summarisation datasets \cite{zhang2019PEGASUS}.
    \item \textbf{T5} is also a Transformer encoder-decoder model which is pre-trained on a multi-task mixture of unsupervised and supervised tasks and unifies all natural language processing tasks into text-to-text format \cite{2020t5}.
    \item \textbf{GPT-Neo}: is an open-sourced GPT3-like model but is pre-trained on the Pile corpus with Masked Autoregressive Language Model. The Pile corpus provides large and diverse text resources for language modelling \cite{gao2020pile}. 
    \item \textbf{Content Selection and Planning} proposes a two-step approach \cite{DBLP:journals/corr/abs-1809-00582}. In the first stage, given a corpus of data records (table-report pairs), the extractor produces a content plan highlighting the values to be mentioned in a specific order. Then in the second stage, an Long Short-Term Memory (LSTM) encoder-decoder model with a copy mechanism is used to generate the report based on the content plan.
\end{itemize}

Table \ref{tab:all_sota_results} shows results by comparing the proposed approach with the SOTA baseline methods on Assay Validation Reports and Toxicology Reports. On Assay Validation Reports, our approach outperforms the SOTA methods by at least an absolute increase of 0.13 on Table Recall, 7.43 on BLEU Extract, 0.05 on ROUGE1. However, GPT-Neo exhibits an exaggerated Table Recall score of 0.86 due to the fact that the model tends to simply copy input tables into the output, hence other scores like BLEU Extract and ROUGE are rather low. On Toxicology Reports, due to the limited amount of real training data, all the baseline methods and the proposed approach are trained using synthetic data. As a result, the proposed approach achieves comparable results as the baseline methods and we tend to attribute this fact to the extremely small size of the test set (44 examples) of Toxicology Reports.

% Results are in Table \ref{table:AVR test} and Table \ref{table:HIL AVR test} for the AVR data and in Table \ref{table:Toxicology_baseline} for the toxicology reports. Overall, for raw AVR and HIL AVR, our approach outperforms the SOTA approaches by a large gap of 0.1 Table Recall, 6.5 BLEU Extract and 0.01 ROUGE1 on average.  GPT-Neo exhibits an exaggerated Table Recall score of 0.86 due to the fact that the model tends to simply copy input tables into the output (hence other scores, like, e.g., BLEU Extract and ROUGE are rather low). 

% For HIL AVR, the Template-based baseline outperforms our method across ROUGE, BLEU and TER. This is due to the nature of the data where lexical content of paragraphs lacks diversity, hence the performance is better with templates. As for the results with Table Recall, which focuses on numerical values rather than lexical part, our method outperforms the Template-based baseline by 0.14.

% \blue{Due to the limited amount of data, all the baselines are trained using synthetic data for toxicology report. Our method achieves similar results as the baselines. We tend to attribute this fact to the extremely small size of the test set (30 examples)} 

% demonstrating the best performance for numerical key values. Differences for the other metrics are less pronounced and, for BLEU Extract, our method is outperformed by PEGASUS marginally. We tend to attribute this fact to the extremely small size of the test set (29 examples).

\subsection*{Ablation Study}

\begin{table}[h]
\renewcommand{\arraystretch}{1.1}
\begin{subtable}[h]{\textwidth}
\centering
\scalebox{0.83}{
\begin{tabular}{|l|c|c|c|c|c|c|c|}
\hline
& Table Recall $\uparrow$ & BLEU Extract $\uparrow$ & ROUGE 1 $\uparrow$ & ROUGE 2 $\uparrow$ & ROUGE L $\uparrow$ & BLEU $\uparrow$ & TER $\downarrow$ \\ 
\hline
Ours & \textbf{0.7245} & 46.92 & \textbf{0.6590} & 0.5304 & 0.5986 & 44.36 & \textbf{0.6455}\\\hline
minus Autocorrector (AC) & 0.7188 & 43.54 & 0.6436 & 0.4991 & 0.5759 & 43.37 & 0.6836 \\ \hline
minus AC, Curriculum Learning (CL) & 0.6966 & 46.49 & 0.6497 & 0.5189 & 0.5899 & 47.85 & 0.6615\\ \hline
minus AC, CL, Synthetic Data & 0.6497 & \textbf{47.61} & \textbf{0.6591} & \textbf{0.5422} & \textbf{0.6029} & \textbf{48.85} & 0.6470 \\ \hline
\end{tabular}}
\caption{\label{table:avr_ablation} Ablation study results on Assay Validation Reports.}
\end{subtable}
\hfill
\begin{subtable}[h]{\textwidth}
\centering
\scalebox{0.9}{
\begin{tabular}{|l|c|c|c|c|c|c|c|}
\hline
& Table Recall $\uparrow$ & BLEU Extract $\uparrow$ & ROUGE 1 $\uparrow$ & ROUGE 2 $\uparrow$ & ROUGE L $\uparrow$ & BLEU $\uparrow$ & TER $\downarrow$ \\ 
\hline
Ours &
\textbf{0.8897} & 29.49 & \textbf{0.9055} & \textbf{0.776} & \textbf{0.9034} & \textbf{65.18} & \textbf{0.264}
% \textbf{0.9051} & 32.1934 & 0.918 & \textbf{0.808} & 0.9133 & 69.8707 & 0.2089
\\\hline
minus Autocorrector (AC) &
0.8694 & \textbf{29.65} & 0.8955 & 0.7719 & 0.8830 & 62.14 & 0.2927
% 0.9023 & \textbf{32.2565} & \textbf{0.9187} & 0.8072 & \textbf{0.9136} & \textbf{70.1443} & \textbf{0.2035} 
\\ \hline
\end{tabular}}
\caption{\label{table:tox_ablation} Ablation study results on Toxicology Reports.}
\end{subtable}
\caption{Ablation study of the proposed approach on Assay Validation Reports and Toxicology Reports to understand the impact of mechanisms: Autocorrector, Curriculum Learning and Synthetic Data. On Assay Validation Reports, the mechanisms together produce higher Table Recall while the other metrics are comparable. On Toxicology Report, due to the limited amount of real training data, the Curriculum Learning is not applicable; Synthetic Data is always applied; and Autocorrector produces comparable results. }
\label{tab:all_ablation_results}
\end{table}

In this section, we investigate the efficiency of the incorporated mechanisms that help our model to be precise while copying crucial table values to the textual output: Synthetic Data, Curriculum Learning and Autocorrector. Table \ref{tab:all_ablation_results} demonstrates the results of the ablation study on Assay Validation Reports and Toxicology Reports which are evaluated on the same test set as the previous section. On Assay Validation Report, Table Recall exhibits a stable improvement from 0.65 to 0.72, benefiting from the addition of those mechanisms, while the BLEU Extract is comparable. This supports our claims on the efficiency of our procedures. As for the other metrics that focus on lexical content (ROUGE, BLEU, TER) our mechanisms tend to maintain them on the same level or slightly decrease them. 

For Toxicology Reports, most metrics exhibit a stable improvement though the BLEU Extract decreases slightly. We hypothesize the decrease of BLEU Extract is caused by the clinical observation section of Toxicology Reports. The clinical observation section usually describes a series of abnormalities, the order of which is not deterministic nor critical, as long as all abnormalities are listed in the text. As a result, though the proposed approach can generate the abnormalities accurately in text, the order may not be the same as that in the target text hence BLEU Extract decreases.

% Tables \ref{table:AE AVR}, \ref{table: AE HIL AVR} and \ref{table:Toxicology AE} report our results. Table Recall exhibits a stable improvement across datasets from 0.65 to 0.72 for AVR, for example, benefiting from the addition of those mechanisms. BLEU Extract tends to increase as well with the exception of the AVR test set. This supports our claims on the efficiency of our procedures. As for the other metrics that focus on lexical content (ROUGE, BLEU) our mechanisms tend to maintain them on the same level or slightly decrease them (e.g., 0.47 ROUGE1 for AVR or decrease of ROUGE1 from 0.63 to 0.62 for HIL AVR). 

% For toxicology data, Table Recall increases by 1\% as well as BLUE score. The overall results support our claims on the efficiency of the procedures again.

%\subsection*{Low-Resource and Light-Weight Model Adaptation via Human-in-the-Loop} 

%\subsection*{Performance Degradation with New Data} 

\begin{table}[h]
\begin{center}
\scalebox{0.83}{
\begin{tabular}{|l|c|c|c|c|c|c|c|}
\hline
& Table Recall $\uparrow$ & BLEU Extract $\uparrow$ & ROUGE 1 $\uparrow$ & ROUGE 2 $\uparrow$ & ROUGE L $\uparrow$ & BLEU $\uparrow$ & TER $\downarrow$ \\ 
\hline
Template & 0.4375 & 24.58 & \textbf{0.6452} & \textbf{0.518} & \textbf{0.5947} & 46.4458 & \textbf{0.6049}\\\hline
Content Selection and Planning & 0.4179 & 10.95 & 0.3264 & 0.1608 & 0.2471 & 10.66 & 1.8715\\\hline
GPT-Neo & \textbf{0.8878} & 24.94 & 0.2074 & 0.0838 & 0.1199 & 2.49 & 1.7256\\\hline
T5 & 0.4966 & 19.90 & 0.3790 & 0.2098 & 0.2610 & 14.39 & 1.3084\\\hline
PEGASUS & 0.4698 & 20.67 & 0.6107 & 0.4856 & 0.5605 & 40.42 & 0.7902 \\\hline
\hline
Ours & \textbf{0.5766} & \textbf{30.37} & 0.6200 &0.5029 & 0.5698 & 39.89 & 0.8641\\\hline
minus Autocorrector (AC) & 0.5728 & 28.20 & 0.6133 & 0.4787 & 0.5513 & 39.80 & 0.8787 \\ \hline
minus AC, Curriculum Learning (CL) & 0.5502 & 27.72 & 0.6056 & 0.4751 & 0.5478 & 39.52 & 0.8426\\ \hline
minus AC, CL, Synthetic & 0.4788 & 24.50 & 0.6328 & 0.5084 & 0.5823 & \textbf{42.66} & 0.7554 \\ \hline
\end{tabular}}
\end{center}
\caption{\label{table:hil_avr_results} Results of comparison with state-of-the-art baseline methods and ablation study on the Assay Validation Reports (AVR) in the Human-in-the-Loop (HIL) scenario. Please note, in these results, the proposed approach and model-based baselines are not fine-tuned by the manually curated data. Overall, though the proposed approach still produces higher Table Recall and BLEU Extract with comparable or better performance on other metrics than baseline and ablated methods, the overall performance decreases on the new data source of AVR for HIL.}
\end{table}

Following on the ablation analysis (Table  \ref{tab:all_ablation_results} (a)), we observe a degradation of performance when the models are applied to the AVR for HIL testing set (Table \ref{table:hil_avr_results}). Note that our approach still outperforms the state-of-the-art methods with at least an absolute increase of 0.08 on Table Recall, 5.43 on BLEU Extract, and shows comparable or better performance on other metrics. Such degradation exemplifies the real-life scenario where new Assay Validation Reports are likely to come from different data sources (e.g. labs, vendors, clinical research organizations). Therefore, the type of input table and the writing style of the reports can be different from historical data.

\subsection*{Model Adaptation utilising Efficient Human-in-the-Loop Protocols} 

To help a model adapt to new data sources efficiently with minimal human effort, we introduce a Human-in-the-Loop (HIL) protocol whereby the model can incorporate a small set of human corrections into its learning processes and thereby create a ``loop'' in which the model constantly learns to improve its capabilities. With the HIL protocol, domain experts who are conversant with the relevant scientific reports are requested to check and correct a sample output automatically generated by the model. The corrections comprise a feedback loop which provides the model with additional training data to provide model adaptation. Ideally, the model should follow the principle of low-resource model adaptation, i.e. relying on a minimal amount of data samples, therefore it is critical that the most rewarding in this regard are selected for human scrutiny.

\begin{figure}[H]
 \centering
 \includegraphics[width=1.0\textwidth]{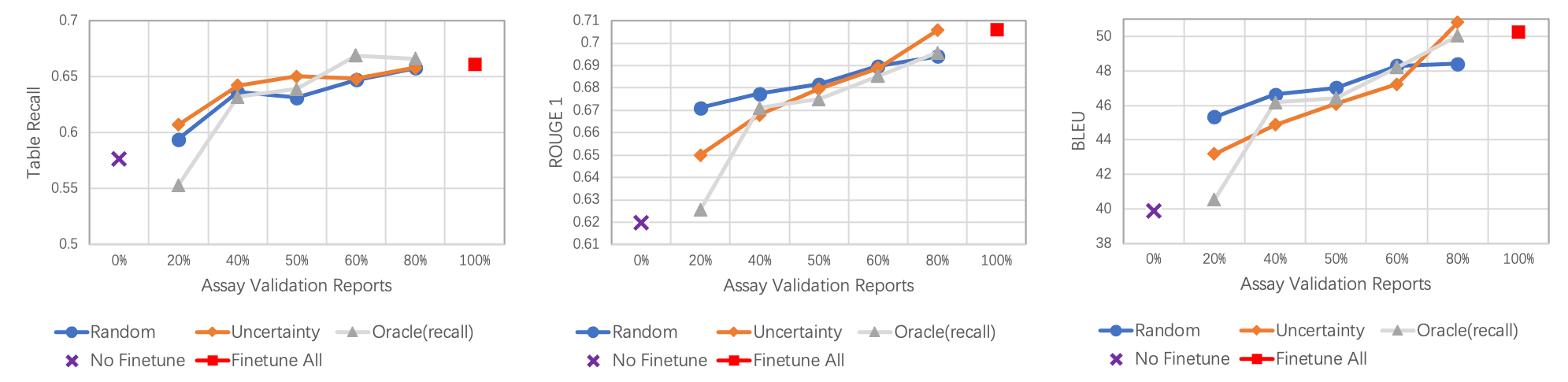}
 \caption{In the low-resource setting with selective supervision, we pick subsets with 0\% (no fine-tuning as presented in Table \ref{table:hil_avr_results}), 20\%, 40\%, 50\%, 60\%, 80\%, and 100\% (full fine-tuning) labelled data to fine-tune and we reports results across Table Recall, ROUGE1 and BLEU. With only 40\% data, the fine-tuned model produces decent performance, and with 80\% data, the model shows similar Table Recall, ROUGE1 and better BLEU than the model that is fully fine-tuned. }\label{fig:selective_supervision}
\end{figure}

In order to avoid the costly prospect of adapting the entire pipeline, i.e., Table Extractor, Text Generator and Autocorrector, we propose light-weight model adaptation paradigm where only the Autocorrector part adapt based on the human-in-the-loop feedback. 
%The Autocorrector is designed to correct the wrong phrases (e.g. numbers, concepts, etc.) in the initial text generated by Table Extractor and Text Generator based on the input tables. 
% An Autocorrector can reduce the workload of human experts to manually post-edit the generated reports. 
% Therefore, 
%In the following experiments where fine-tuning is required, only Autocorrector is fine-tuned.

%As manually curated data is expensive in real-world practice, when only limited gold data is available, i.e., in the low-resource setting, 
Figure \ref{fig:selective_supervision} compares three data sample selection strategies for Active Learning, namely random sampling, uncertainty-based sampling and oracle sampling, as applied to AVR for HIL. Overall, in the scenario with only 40\% of curated data, the model shows reasonable performance across strategies. In the scenario with 80\% of curated data, the model shows similar (in terms of Table Recall, ROUGE1) or even better performance (in terms of BLEU) than the model that is fully fine-tuned. The uncertainty-based sampling strategy tends to be particularly beneficial in terms of improving Table Recall in the scenario where less than 50\% of curated data is available. 

Table~\ref{tab:result_table} shows an example of the model output (predicted text) with reliably predicted relevant values contrasted by the human-generated text.

\newcommand{\tabincell}[2]{\begin{tabular}{@{}#1@{}}#2\end{tabular}}
\begin{table}[htbp]
\centering
\caption{\label{tab:result_table} Example of Assay Validation Report for Human-in-the-Loop. Side-by-side comparison of model output and human-generated text. Relevant values are highlighted in bold. For proprietary reason, the values in the examples are dummy for demonstration purpose only.}
\begin{tabular}{| p{0.42\columnwidth}|p{0.42\columnwidth}|} 
\hline
Predicted text & Human-generated text \\
\hline
 The effect of \textbf{ProcessH} on the detection of \textbf{ConditionA} was evaluated in run RunID92 by analyzing blanks ( \textbf{GroupH} \textbf{1}, \textbf{4}, and \textbf{7} ), low - level ( \textbf{GroupH} \textbf{3}, \textbf{6}, and \textbf{9} at \textbf{300} ng/mL ), and high - level ( \textbf{\textbf{GroupH}} \textbf{2}, \textbf{5}, and \textbf{8} at \textbf{1000} ng/mL ) controls, prepared in \textbf{0} \%, \textbf{2} \%, or \textbf{10} \% ProcessH whole blood, respectively. As presented in \textbf{Table7B}, there was no effect from \textbf{ProcessH} on the detection of low - level \textbf{ConditionA} and high - level \textbf{ConditionA} in \textbf{2} \% and \textbf{10} \% \textbf{ProcessH} whole blood, and no cross - reactivity was observed in negative samples.& The effect of \textbf{ProcessH} on the detection of \textbf{ConditionA} was evaluated in runs \textbf{RunID8} and \textbf{RunID9} by analyzing blanks ( \textbf{GroupH} \textbf{1} , \textbf{4} , and \textbf{7} ) , low - level ( \textbf{GroupH} \textbf{2} , \textbf{5} , and \textbf{8} at \textbf{300} ng/mL ) , and high - level ( \textbf{GroupH} \textbf{3} , \textbf{6} , and \textbf{9} at \textbf{1000} ng/mL , each pre - spiked with 250 $\mu$g/mL of DrugA ) positive controls , prepared in \textbf{0} \% , \textbf{2} \% , or \textbf{10} \% ProcessH human whole blood . As presented in \textbf{Table7B} , there was no effect from \textbf{ProcessH} on the detection of low - and high - level \textbf{ConditionA} in \textbf{2} \% and \textbf{10} \% \textbf{ProcessH} whole blood , and no cross - reactivity was observed in negative samples .\\
%\hline

%Worst & The fixed specificity assay cut point ( CP.V ) was generated based on percentile ( e.g.. 95th percentile ) of the pooled data ( \textbf{Table1H} and Appendix C ) Please select appropriate cut point based on scientific purpose. & From the intra - assay result , \textbf{PTC} 's titer \% \textbf{CV} for \textbf{RUN} \textbf{RunID1} was 0.2 \% . Data for these runs are presented in \textbf{Table1H} .\\
\hline
\end{tabular}
\end{table}

\section*{Discussion}

We present an approach to generate extractive-abstractive paragraphs summarising scientific medical tables. Our solution presents a unique combination of the state-of-the-art text generation means (Copy Mechanisms, Curriculum Learning, Synthetic Data and Autocorrector) to ensure the creation of models that address the crucial challenge of the scientific Table2Text: accurate summarisation of key tabular values. To our knowledge, such type of Table2Text generation of scientific medical text has not been tackled before.

We propose an extended approach to generate synthetic examples of table-text tuples to address the extreme data sparsity in the domain (e.g., the 43 Toxicology Reports training pairs we use in our study).
Combined with the Copy Mechanisms that act like pointers, our data augmentation procedure promotes precision of copying relevant values from inputs and increase the model's potential to be interpretable. 

Moreover, to address the need to rapidly adapt the model to new types of data we use the Active Learning (AL) techniques in the Human-in-the-Loop protocol that ensure the minimal human correction effort. Then we choose not to fine-tune large models in a costly way but rather perform light-weight updates to the Autocorrector model built on top of the main model.
 
There are several directions to take our work further. The underlying deep learning methodology of the Extractor and the Generator could be changed to use methods that allow longer tabular inputs (e.g., Big Bird~\cite{bigbird}).

Finally, we have investigated only two types of text, namely Assay Validation Reports and Toxicology Reports. Looking forward, a more universal approach to generate different types of reports with one single model would contribute to the model generalisability and further address the data sparsity.\cite{johnson-etal-2017-googles}

% \red{TODO for Michael: If you could help with (1) discussing impact and values of this study to pre-clinical and clinical study teams and (2) (pre-)clinical interpretation of the model's performance, that would be great to strengthen the results and discussion.}

In conclusion, the proposed Table-to-Text Generation model makes impacts in several aspects of AI-Assisted Report Writing. With the precise value extraction from tables and the concise text generation for drafting a report, it dramatically (1) eliminates the manual annotation need on prospective reports, (2) increases data traceability and reduces human errors from copy and pasting across different tables or database, and (3) ultimately speeds up regulatory filing for pre-clinical and clinical study teams.

\section*{Methods}

\subsection*{Data Pre-processing} 

In the table pre-processing step, we flatten the input tables row by row into a sequence of tokens. To reduce the token count, we reduce tokens in table rows if they re-appear at the same position in at least two consecutive table rows. We also limit the number of table rows and the number of tokens per row to accommodate 85\% of corpus-level matches of table values to reports. We prepend each table with its title. Special cell markup (e.g., asterisks indicating a remarkable result) is propagated to respective column and row titles. 

Besides, we also observe the writing of Toxicology Reports typically follows certain templates by rules. For example, if a group has multiple individuals, only the mean value of individuals is reported instead of listing the results of all individuals. Similarly, if a control group is available, the difference between the experimental group and the control group is typically reported. Therefore, we implement such rules in table pre-processing to filter the input information for the model.

\subsection*{Architecture} 

\begin{figure}[htbp]
\begin{floatrow}
\ffigbox{\center{\includegraphics[width=.99\linewidth]{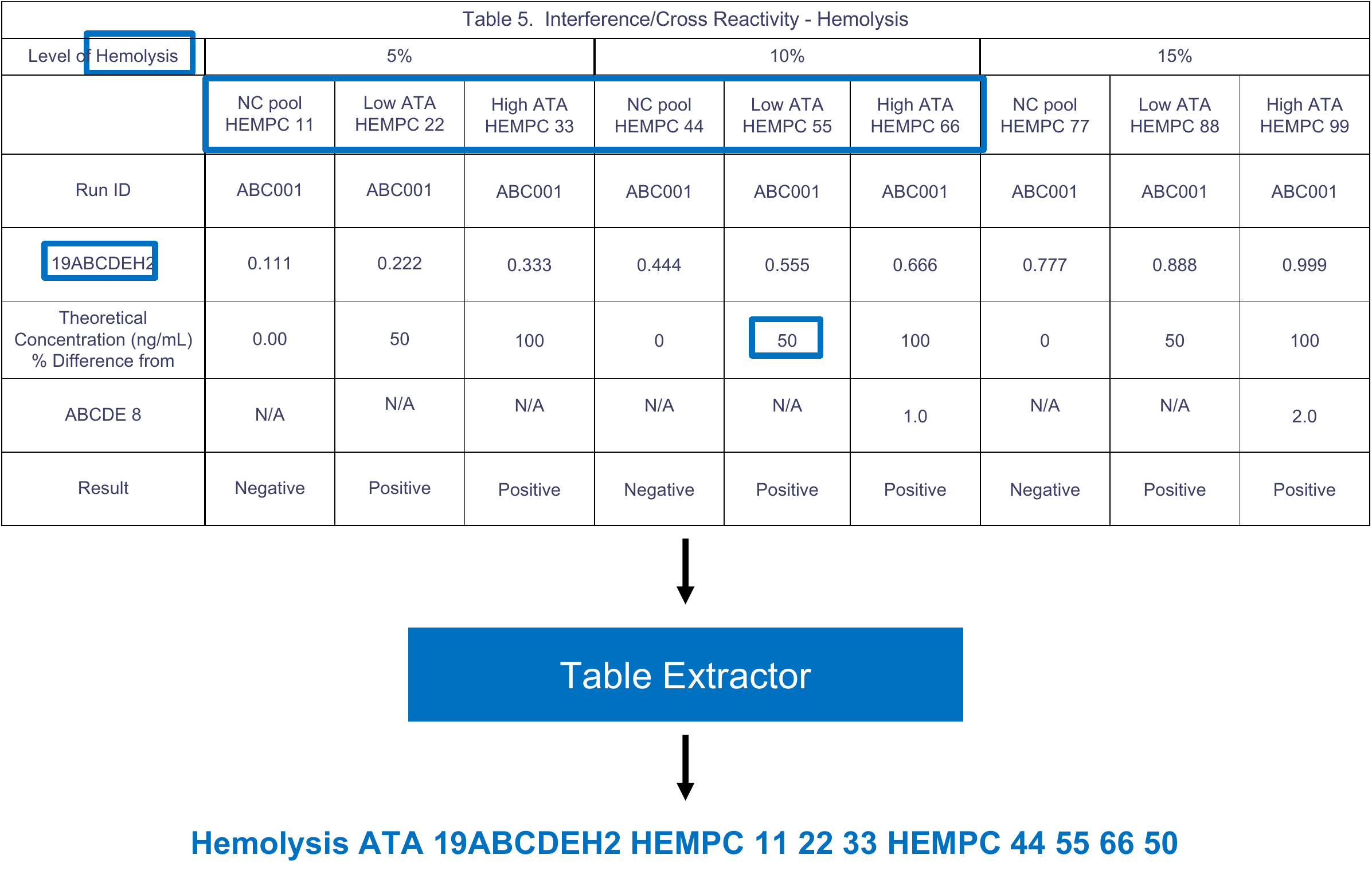}}}
{\caption{The Table Extractor takes full tables (one or more) as the input and generates the extract of key tabular values. The extract reflects the values from the tables in the count and order they should appear in the target report. \label{fig:extractor}}}
\ffigbox{
\center{\includegraphics[width=.99\linewidth]{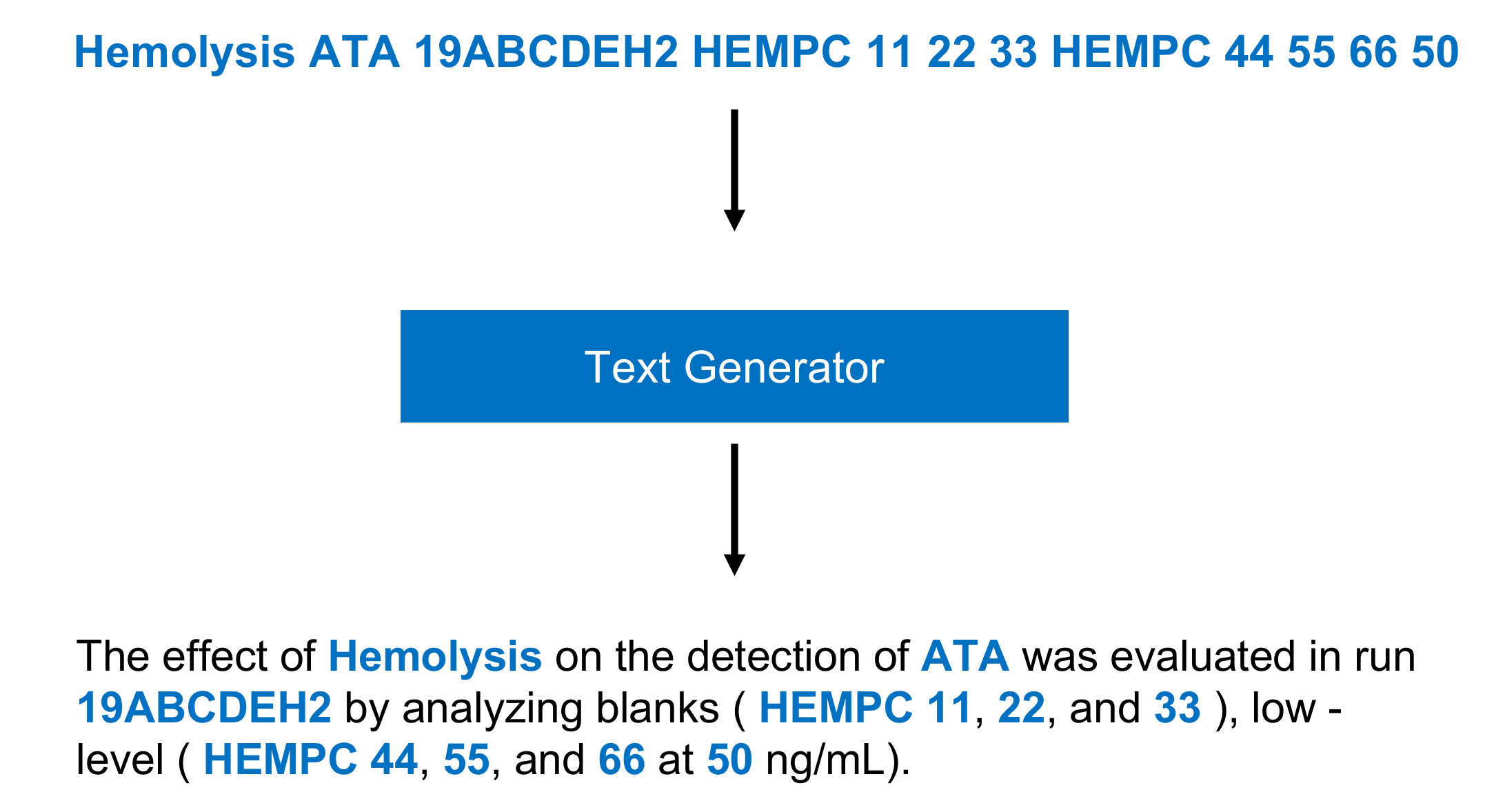}}}
{\caption{The values extracted from Table Extractor is fed into the Text Generator to generate the output text. \label{fig:generator}}}
\end{floatrow}
\end{figure}

Following best practices in the domain, we apply a two-step approach that consists of: 

\begin{itemize}
    \item \textbf{Table Extractor} (Figure~\ref{fig:extractor}) takes full table(s) as the input and generates the extract of key values. The table extract contains values from the table(s) in the order that they should eventually appear in the target text. It can also contain repetitions if the values are repeated in the target text. The table extract usually contains biomedical concepts (like capitalised words or their groups that are names of drug and disease), names of runs, abbreviations, float and integer numerical values, as well as asterisks that mark words and numbers. Asterisks mark important information with respect to outstanding events and are essential for correct text generation. The table extracts do not contain other characters like punctuation marks (e.g., \%) or common words thus focusing on only important tabular values.
    \item \textbf{Text Generator} (Figure~\ref{fig:generator}) takes the table extract (from one or multiple tables) as generated by the Table Extractor and generates the respective textual paragraph. Before the table extract is fed into the Text Generator, it is first prepended with the titles of all the tables and appended with the last three rows of each full table. This information has been empirically found beneficial for text generation.  
\end{itemize}

Both Table Extractor and Text Generator are implemented by Transformer Encoder-Decoder architecture. The decoder is a conditional language model which generates a new word at each time step by taking into account all previously generated words as well as the information provided by the encoder. Both models are equipped with a Copy Mechanism.

% In this work, the input to the Extractor encoder is the original table and the output of the decoder is the table extract. The input to the Generator encoder is the table extract and the output is the report. Each model is equipped with a CopyMechanism.

In its nutshell, the Copy Mechanism follows the practice of the previous study \cite{ive-etal-2019-deep}. We first obtain an attention distribution over the encoder representations at each decoder time step. Then a switching function is incorporated to decide whether to generate a word by the decoder or to copy an important word directly from the input (which is input table tokens for Table Extractor or the table extract for the Text Generator).

Following the best practice in the domain~\cite{zhang2019PEGASUS}, both models are initialised with weights that have been pre-trained to capture the language of the scientific medical literature. These weights are then fine-tuned for our datasets. In the low-resource condition, this fine-tuning needs to be performed with synthetic data to ensure optimal results (see~\ref{ssec:synt-data}). After the fine-tuning by the entire dataset, the model is further fine-tuned on the most difficult examples in the Curriculum Learning setting~\cite{see-etal-2017-get}. For the Assay Validation Reports, the most of difficult examples are those that require copying values from the middle and end of long tables. 

\subsection*{Synthetic Data} \label{ssec:synt-data}  

As mentioned above, the Table2Text generation task assumes describing certain values from tables with textual narratives and the values that come from tables and are mentioned in the text can be matched. We take a different combination of those matched values to condition the creation of synthetic training examples. Following best domain practice ~\cite{chang-etal-2021-neural}, the combinations of key values are picked up randomly as soon as positional value type is respected. For example, a key tabular value ``Hemolysis'' could be replaced by a word/group of words used for a similar slot type (e.g., ``Lipolysis'') based on the dictionary we extract from the original training data. Overall, we define the following value types: integer values, float values, run IDs, string values and table IDs. String values include drug and disease and other biomedical abbreviations (e.g., ADA for anti-drug antibody, CV for the coefficient of variation).
Variations of integer, float and alphanumerical run ID values are created by slightly modifying them, for example, by taking the value close to the original (e.g., 0.65 instead of 0.6), or by randomising the order for alphanumeric values (e.g., ``ABC123'' modified to ``A1BC23''). 

% The amount of modification is fine-tuned with respect to each dataset. 

Respective synthetic training examples (both tables and reports) are created by replacing the matched values with those randomly selected key values. 
The combination of the Copy Mechanism and the synthetic training examples where only the matched values change allows the model to focus on the precision of the copied values. Also, the Copy Mechanism acts like a pointer to the relevant input values and is transparent to model users.

\subsection*{Automatic Corrector (Autocorrector)} 

\begin{figure}[htbp]
\begin{floatrow}
\ffigbox{\center{\includegraphics[width=.6\linewidth]{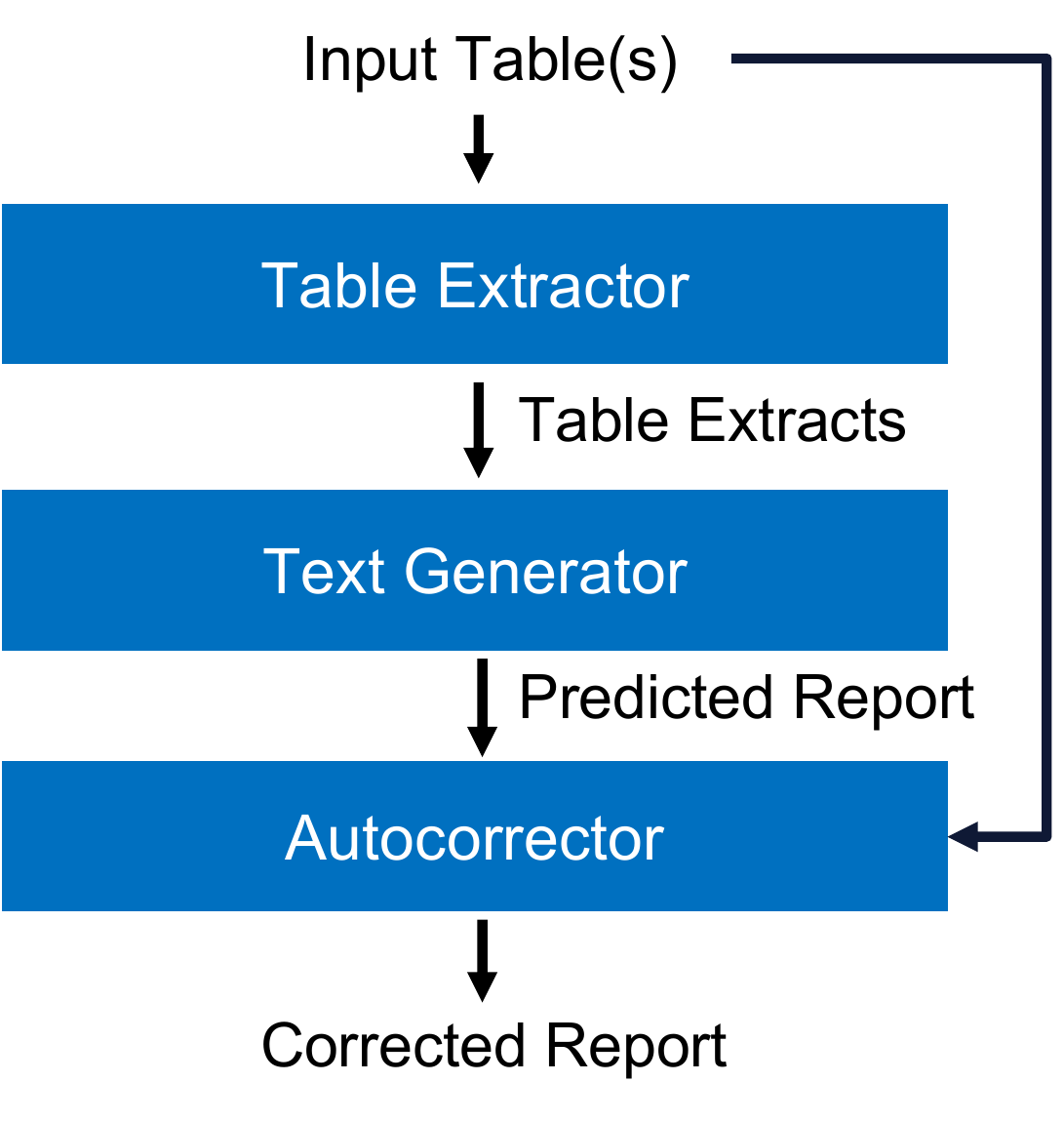}}}
{\caption {The pipeline with Table Extractor, Text Generator and Autocorrector. The Autocorrector allows light-weight updates to the entire pipeline and shows improved performance by correcting tabular values in text. \label{fig:ape-flow}}}
\ffigbox{
\center{\includegraphics[width=.99\linewidth]{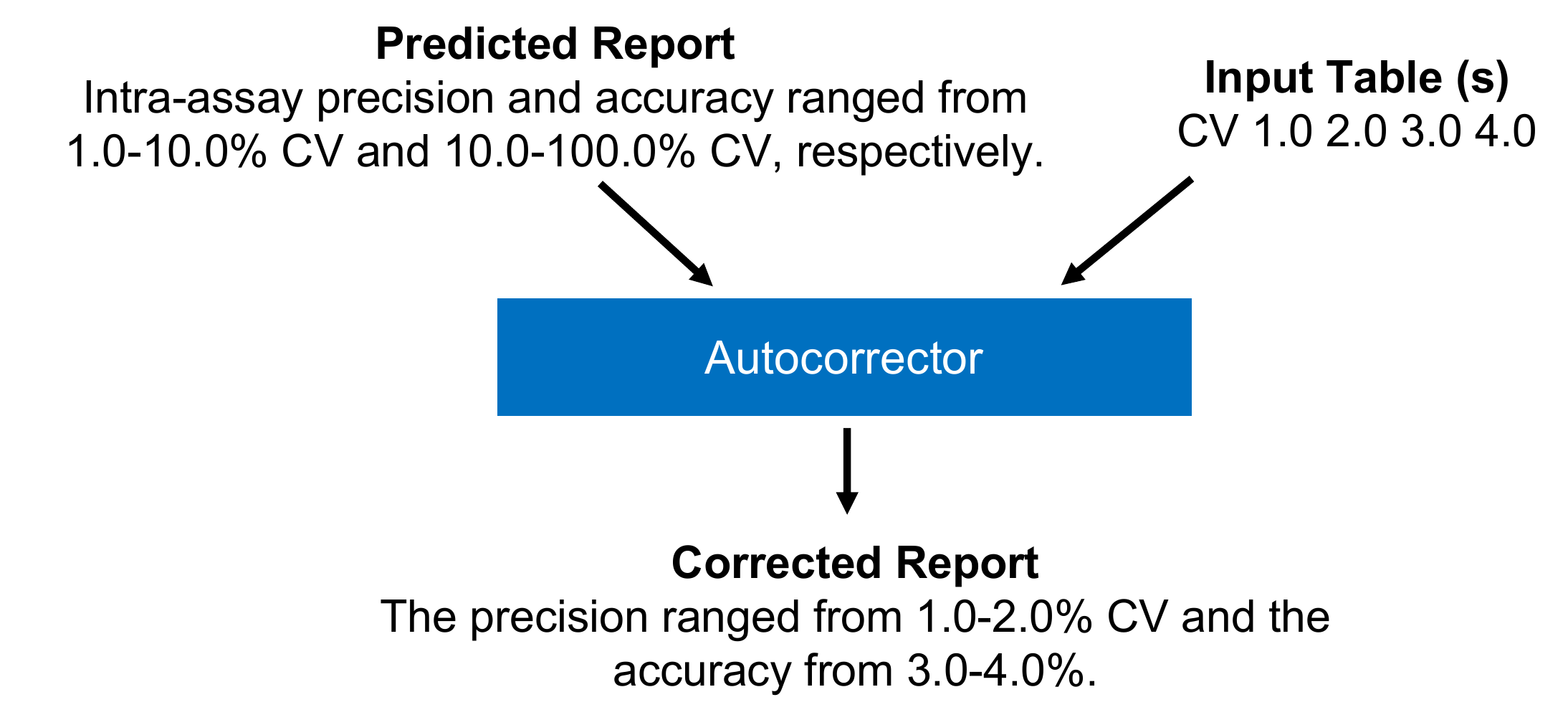}}}
{\caption {The Autocorrector by Automatic Post-Editing (APE) takes the initially predicted report and the input table(s) to produce a new report by correcting values. \label{fig:ape-example}}}
\end{floatrow}
\end{figure}

To enable light-weight updates to the two-step architecture with Table Extractor and Text Generator, we add an Autocorrector model following the Text Generator (see Figure~\ref{fig:ape-flow}). The Autocorrector is also a Transformer Encoder-Decoder. The inputs to Autocorrector are the initially generated text by Text Generator and the original input table (full table but not the table extract, see Figure~\ref{fig:ape-example}) and the Autocorrector corrects the tabular values in the report accordingly.
% We also reverse the order of values in those tables that promotes correction using the values coming from the middle and end of tables (more difficult cases of copying).
The Autocorrector model follows the best practices of Automatic Post-Editing (APE)~\cite{chatterjee-EtAl:2020:WMT} that seeks to reduce the burden of human post-editors and automatically corrects errors in machine-translated outputs.

\subsection*{Human-in-the-Loop Protocol} 

\begin{figure}[hbt!]
 \centering
\includegraphics[width=0.99\textwidth]{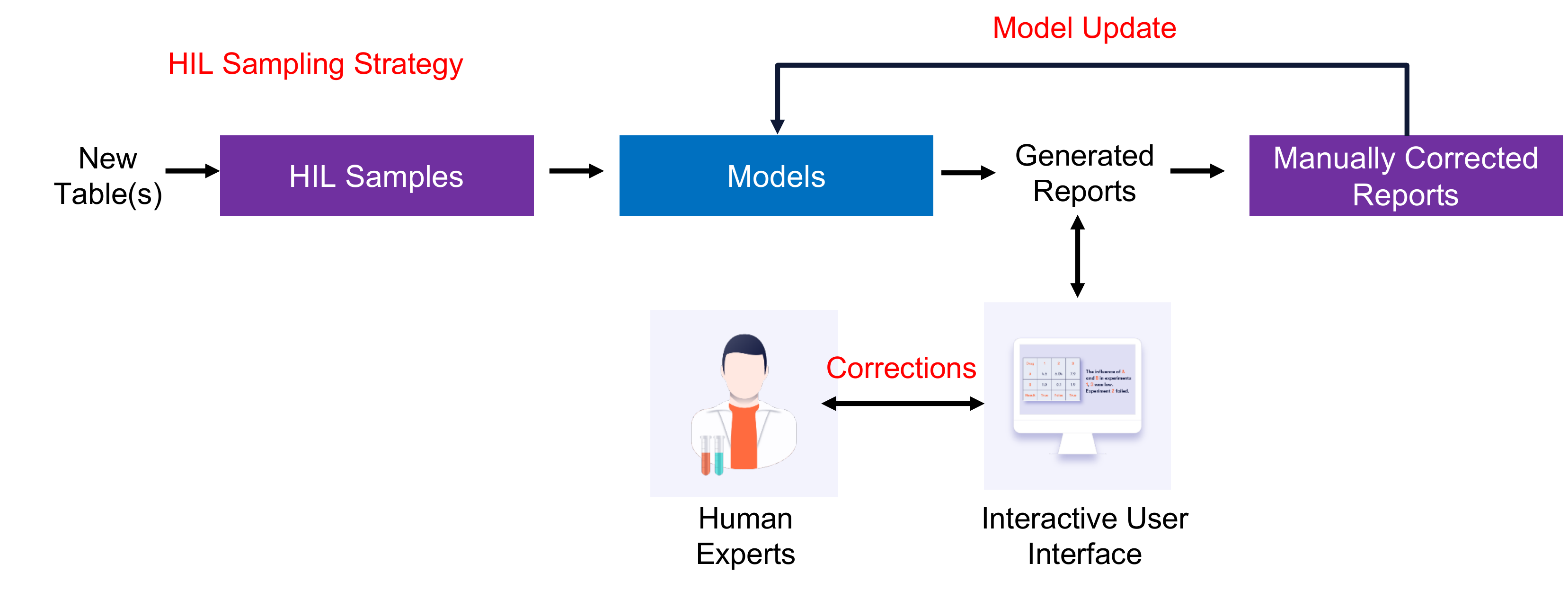}
 \caption{Human-in-the-Loop (HIL) protocol for on-the-fly Table2Text model adaptation using the Autocorrector model: (1) selection of a pool of the most beneficial samples to be annotated; (2) generation of reports with the Extractor-Generator model; (3) correction of generated reports by human experts; (4) learning of the Autocorrector to rewrite initial outputs according to the human feedback. }\label{fig:hil}
\end{figure}

Our Human-in-the-Loop (HIL) protocol for on-the-fly model adaptation is in Figure~\ref{fig:hil}. The protocol is composed of the following steps: (1) selection of a pool of the most beneficial samples from unlabelled samples according to a pre-set selection strategy; (2) generation of reports with the two-step approach (not involving the Autocorrector); (3) correction of generated reports by human experts via a designated interface; (4) learning of the Autocorrector to rewrite the two-step approach outputs according to the human feedback. 

The human interface developed for the protocol enables the selection and the visualisation of both the original table and the generated text. Table extracts were not displayed to the human experts to decrease the cognitive load. Human annotators received instructions to perform as few corrections as possible as long as the text is clinically valid. 

As the beneficial sample selection strategy, each time we pick a certain percentage of training examples according to one of the following strategies: (1) \textbf{Random sampling}: the samples are selected at random; (2) \textbf{Uncertainty-based sampling}: the entropy score is computed based on the predicted probability of generated tokens to measure the uncertainty of the Autocorrector model for each sample, and then the samples that have the highest entropy score are selected; (3) \textbf{Oracle sampling}: we measure the Table Recall between the generated text by the Autocorrector and the target text and the samples with lowest Table Recall are selected.

\subsection*{Implementation Details}

The code is developed by Python 3.8 and Pytorch~\cite{NEURIPS2019_bdbca288}. Our pre-processing pipeline uses the spaCy toolkit \footnote{spaCy: \url{https://spacy.io/}}. 

% Assay Validation tables were limited to table row threshold of 20 tokens and the overall table token threshold of 25 tokens.

All the three models (Table Extractor, Text Generator and Autocorrector) are the same state-of-the-art Transformer-based PEGASUS text summarisation architecture~\cite{zhang2019PEGASUS} and are initialised by the pre-trained weights \texttt{PEGASUS-pubmed} setup from the transformers library \cite{wolf-etal-2020-transformers}. For fine-tuning, we use the Adafactor~\cite{adafactor} optimiser and train each model for 50,000 steps with  a batch size of 4. For the synthetic data generation, we generate 1,000 examples per each real training example. For the Curriculum Learning setting, we fine-tune the Table Extractor model for 20,000 steps further with 140 difficult examples (140,000 after syntheitc data generation) that imply copying from the middle and end of tables.

% Instead of using synthetic data, our baseline models are only trained on real data unless they are specified. 

For the PEGASUS baseline model \cite{zhang2019PEGASUS}, the setup is the same as our models and we fine-tune PEGASUS for 30,000 steps with the batch size of 4. For the T5 baseline model, we use \texttt{t5-large} setup as the pre-trained weights and fine-tune T5 for 30,000 steps. We update the length penalty to 1.2 to encourage the model to generate longer sequences. For GPT-Neo baseline model, we use the \textit{GPT-Neo-125M} setup from EleutherAI\cite{gao2020pile}. To fine-tune the GPT-Neo model for text generation, we use the teacher forcing method which uses ground truth from a prior time step as input and trains the model for 30,000 steps.
The content selection and planning baseline is developed based on a third-party codebase\cite{DBLP:journals/corr/abs-1809-00582}. The gold-standard content plan is constructed from the matches between the input tables and target reports. 

% \begin{table}[H]
% 	\centering
% 	\renewcommand{\arraystretch}{1.1}
% 	\begin{tabular}{|c|c|} 
% 	    \hline
% 	    \multicolumn{2}{|c|}{Training for Content Selection and Planning} \\ \hline
% 		Optimiser               & AdaGrad \\
% 		AdaGrad accumulator init & 0.1 \\
% 		Learning rate           & 0.15  \\
% 		learning rate decay   & 0.97\\
% 		Start decay at        &  4 \\
% 		Batch size              & 5  \\
% 		\hline

% 	\end{tabular}
% 	\caption{Hyper-parameters for training are decided empirically on the validation set.}
% 	\label{tab:hyperparam-erac}
% \end{table}

\section*{Code and Data Availability}

Code and data could not be shared publicly due to their proprietary nature.

% \section*{Acknowledgements}

% We would like to thank scientific experts (\red{TODO: names}) from Genentech Inc. for their feedback and comments throughout this research. 

\section*{Author contributions statement}

JI conceived the architecture and the automatic corrector approaches. JZ conceived the experiments. HW prepared and processed the datasets. HW curated the datasets with expertise. JZ and TL conducted the experiments. JI, JZ and TL analysed the results. JI, VG, BC and YG reviewed the research and manuscript. All authors approved the manuscript.
 
\section*{Additional information}

\textbf{Competing interests}: The work is under collaboration between Genentech Inc and Pangaea Data Limited.
% , Imperial College London, Queen Mary University of London and Hong Kong Baptist University.

\bibliography{sample}

\begin{thebibliography}{10}
\urlstyle{rm}
\expandafter\ifx\csname url\endcsname\relax
  \def\url#1{\texttt{#1}}\fi
\expandafter\ifx\csname urlprefix\endcsname\relax\def\urlprefix{URL }\fi
\expandafter\ifx\csname doiprefix\endcsname\relax\def\doiprefix{DOI: }\fi
\providecommand{\bibinfo}[2]{#2}
\providecommand{\eprint}[2][]{\url{#2}}

\bibitem{chang-etal-2021-neural}
\bibinfo{author}{Chang, E.}, \bibinfo{author}{Shen, X.}, \bibinfo{author}{Zhu,
  D.}, \bibinfo{author}{Demberg, V.} \& \bibinfo{author}{Su, H.}
\newblock \bibinfo{title}{Neural data-to-text generation with {LM}-based text
  augmentation}.
\newblock In \emph{\bibinfo{booktitle}{Proceedings of the 16th Conference of
  the European Chapter of the Association for Computational Linguistics: Main
  Volume}}, \bibinfo{pages}{758--768} (\bibinfo{publisher}{Association for
  Computational Linguistics}, \bibinfo{address}{Online}, \bibinfo{year}{2021}).

\bibitem{chang-etal-2021-order}
\bibinfo{author}{Chang, E.}, \bibinfo{author}{Yeh, H.-S.} \&
  \bibinfo{author}{Demberg, V.}
\newblock \bibinfo{title}{Does the order of training samples matter? improving
  neural data-to-text generation with curriculum learning}.
\newblock In \emph{\bibinfo{booktitle}{Proceedings of the 16th Conference of
  the European Chapter of the Association for Computational Linguistics: Main
  Volume}}, \bibinfo{pages}{727--733} (\bibinfo{publisher}{Association for
  Computational Linguistics}, \bibinfo{address}{Online}, \bibinfo{year}{2021}).

\bibitem{puduppully2021datatotext}
\bibinfo{author}{Puduppully, R.} \& \bibinfo{author}{Lapata, M.}
\newblock \bibinfo{title}{{Data-to-text Generation with Macro Planning}}
  (\bibinfo{year}{2021}).
\newblock \eprint{2102.02723}.

\bibitem{wang-etal-2020-towards}
\bibinfo{author}{Wang, Z.}, \bibinfo{author}{Wang, X.}, \bibinfo{author}{An,
  B.}, \bibinfo{author}{Yu, D.} \& \bibinfo{author}{Chen, C.}
\newblock \bibinfo{title}{Towards faithful neural table-to-text generation with
  content-matching constraints}.
\newblock In \emph{\bibinfo{booktitle}{Proceedings of the 58th Annual Meeting
  of the Association for Computational Linguistics}},
  \bibinfo{pages}{1072--1086}, \doiprefix\url{10.18653/v1/2020.acl-main.101}
  (\bibinfo{publisher}{Association for Computational Linguistics},
  \bibinfo{address}{Online}, \bibinfo{year}{2020}).

\bibitem{gong-etal-2020-tablegpt}
\bibinfo{author}{Gong, H.} \emph{et~al.}
\newblock \bibinfo{title}{{T}able{GPT}: Few-shot table-to-text generation with
  table structure reconstruction and content matching}.
\newblock In \emph{\bibinfo{booktitle}{Proceedings of the 28th International
  Conference on Computational Linguistics}}, \bibinfo{pages}{1978--1988},
  \doiprefix\url{10.18653/v1/2020.coling-main.179}
  (\bibinfo{publisher}{International Committee on Computational Linguistics},
  \bibinfo{address}{Barcelona, Spain (Online)}, \bibinfo{year}{2020}).

\bibitem{chen-etal-2020-shot}
\bibinfo{author}{Chen, Z.}, \bibinfo{author}{Eavani, H.},
  \bibinfo{author}{Chen, W.}, \bibinfo{author}{Liu, Y.} \&
  \bibinfo{author}{Wang, W.~Y.}
\newblock \bibinfo{title}{{Few-Shot {\{}NLG{\}} with Pre-Trained Language
  Model}}.
\newblock In \emph{\bibinfo{booktitle}{Proceedings of the 58th Annual Meeting
  of the Association for Computational Linguistics}},
  \bibinfo{pages}{183--190}, \doiprefix\url{10.18653/v1/2020.acl-main.18}
  (\bibinfo{publisher}{Association for Computational Linguistics},
  \bibinfo{address}{Online}, \bibinfo{year}{2020}).

\bibitem{vasvani:nips:2018}
\bibinfo{author}{Vaswani, A.} \emph{et~al.}
\newblock \bibinfo{title}{Attention is all you need}.
\newblock In \emph{\bibinfo{booktitle}{Advances in Neural Information
  Processing Systems 30}}, \bibinfo{pages}{5998--6008} (\bibinfo{year}{2017}).

\bibitem{Pauws2019}
\bibinfo{author}{Pauws, S.}, \bibinfo{author}{Gatt, A.},
  \bibinfo{author}{Krahmer, E.} \& \bibinfo{author}{Reiter, E.}
\newblock \emph{\bibinfo{title}{{Making Effective Use of Healthcare Data Using
  Data-to-Text Technology}}}, \bibinfo{pages}{119--145}
  (\bibinfo{publisher}{Springer International Publishing},
  \bibinfo{year}{2019}).

\bibitem{see-etal-2017-get}
\bibinfo{author}{See, A.}, \bibinfo{author}{Liu, P.~J.} \&
  \bibinfo{author}{Manning, C.~D.}
\newblock \bibinfo{title}{Get to the point: Summarization with
  pointer-generator networks}.
\newblock In \emph{\bibinfo{booktitle}{Proceedings of the 55th Annual Meeting
  of the Association for Computational Linguistics (Volume 1: Long Papers)}},
  \bibinfo{pages}{1073--1083}, \doiprefix\url{10.18653/v1/P17-1099}
  (\bibinfo{publisher}{Association for Computational Linguistics},
  \bibinfo{address}{Vancouver, Canada}, \bibinfo{year}{2017}).

\bibitem{vinyals2015pointer}
\bibinfo{author}{Vinyals, O.}, \bibinfo{author}{Fortunato, M.} \&
  \bibinfo{author}{Jaitly, N.}
\newblock \bibinfo{title}{Pointer networks}.
\newblock In \emph{\bibinfo{booktitle}{Advances in Neural Information
  Processing Systems}}, \bibinfo{pages}{2692--2700} (\bibinfo{year}{2015}).

\bibitem{zhang2019PEGASUS}
\bibinfo{author}{Zhang, J.}, \bibinfo{author}{Zhao, Y.},
  \bibinfo{author}{Saleh, M.} \& \bibinfo{author}{Liu, P.~J.}
\newblock \bibinfo{title}{Pegasus: Pre-training with extracted gap-sentences
  for abstractive summarization} (\bibinfo{year}{2019}).
\newblock \eprint{1912.08777}.

\bibitem{activelearning}
\bibinfo{author}{Settles, B.}
\newblock \bibinfo{title}{Active learning literature survey}.
\newblock \bibinfo{type}{Computer Sciences Technical Report}
  \bibinfo{number}{1648}, \bibinfo{institution}{University of
  Wisconsin--Madison} (\bibinfo{year}{2009}).

\bibitem{chatterjee-EtAl:2020:WMT}
\bibinfo{author}{Chatterjee, R.}, \bibinfo{author}{Freitag, M.},
  \bibinfo{author}{Negri, M.} \& \bibinfo{author}{Turchi, M.}
\newblock \bibinfo{title}{Findings of the wmt 2020 shared task on automatic
  post-editing}.
\newblock In \emph{\bibinfo{booktitle}{Proceedings of the Fifth Conference on
  Machine Translation}}, \bibinfo{pages}{646--659}
  (\bibinfo{publisher}{Association for Computational Linguistics},
  \bibinfo{address}{Online}, \bibinfo{year}{2020}).

\bibitem{parton-etal-2012-automatic}
\bibinfo{author}{Parton, K.}, \bibinfo{author}{Habash, N.},
  \bibinfo{author}{McKeown, K.}, \bibinfo{author}{Iglesias, G.} \&
  \bibinfo{author}{de~Gispert, A.}
\newblock \bibinfo{title}{Can automatic post-editing make {MT} more
  meaningful}.
\newblock In \emph{\bibinfo{booktitle}{Proceedings of the 16th Annual
  conference of the European Association for Machine Translation}},
  \bibinfo{pages}{111--118} (\bibinfo{publisher}{European Association for
  Machine Translation}, \bibinfo{address}{Trento, Italy},
  \bibinfo{year}{2012}).

\bibitem{Papieni02bleu}
\bibinfo{author}{Papineni, K.}, \bibinfo{author}{Roukos, S.},
  \bibinfo{author}{Ward, T.} \& \bibinfo{author}{Zhu, W.-J.}
\newblock \bibinfo{title}{{BLEU: a Method for Automatic Evaluation of Machine
  Translation}}.
\newblock In \emph{\bibinfo{booktitle}{Proceedings of 40th Annual Meeting of
  the Association for Computational Linguistics}}, \bibinfo{pages}{311--318}
  (\bibinfo{year}{2002}).

\bibitem{Lin:2004}
\bibinfo{author}{Lin, C.-Y.}
\newblock \bibinfo{title}{{ROUGE}: A package for automatic evaluation of
  summaries}.
\newblock In \emph{\bibinfo{booktitle}{ACL workshop on Text Summarization
  Branches Out}} (\bibinfo{year}{2004}).

\bibitem{snover-AMTA-2006}
\bibinfo{author}{Snover, M.}, \bibinfo{author}{Dorr, B.},
  \bibinfo{author}{Schwartz, R.}, \bibinfo{author}{Micciulla, L.} \&
  \bibinfo{author}{Makhoul, J.}
\newblock \bibinfo{title}{A study of translation edit rate with targeted human
  annotation}.
\newblock In \emph{\bibinfo{booktitle}{Proceedings of Association for Machine
  Translation in the Americas}}, \bibinfo{pages}{223--231}
  (\bibinfo{year}{2006}).

\bibitem{2020t5}
\bibinfo{author}{Raffel, C.} \emph{et~al.}
\newblock \bibinfo{journal}{\bibinfo{title}{Exploring the limits of transfer
  learning with a unified text-to-text transformer}}.
\newblock {\emph{\JournalTitle{Journal of Machine Learning Research}}}
  \textbf{\bibinfo{volume}{21}}, \bibinfo{pages}{1--67} (\bibinfo{year}{2020}).

\bibitem{gao2020pile}
\bibinfo{author}{Gao, L.} \emph{et~al.}
\newblock \bibinfo{journal}{\bibinfo{title}{The pile: An 800gb dataset of
  diverse text for language modeling}}.
\newblock {\emph{\JournalTitle{arXiv preprint arXiv:2101.00027}}}
  (\bibinfo{year}{2020}).

\bibitem{DBLP:journals/corr/abs-1809-00582}
\bibinfo{author}{Puduppully, R.}, \bibinfo{author}{Dong, L.} \&
  \bibinfo{author}{Lapata, M.}
\newblock \bibinfo{title}{Data-to-text generation with content selection and
  planning}.
\newblock In \emph{\bibinfo{booktitle}{Proceedings of the 33rd AAAI Conference
  on Artificial Intelligence}} (\bibinfo{address}{Honolulu, Hawaii},
  \bibinfo{year}{2019}).

\bibitem{bigbird}
\bibinfo{author}{Zaheer, M.} \emph{et~al.}
\newblock \bibinfo{title}{Big bird: Transformers for longer sequences}.
\newblock In \bibinfo{editor}{Larochelle, H.}, \bibinfo{editor}{Ranzato, M.},
  \bibinfo{editor}{Hadsell, R.}, \bibinfo{editor}{Balcan, M.~F.} \&
  \bibinfo{editor}{Lin, H.} (eds.) \emph{\bibinfo{booktitle}{Advances in Neural
  Information Processing Systems}}, vol.~\bibinfo{volume}{33},
  \bibinfo{pages}{17283--17297} (\bibinfo{publisher}{Curran Associates, Inc.},
  \bibinfo{year}{2020}).

\bibitem{johnson-etal-2017-googles}
\bibinfo{author}{Johnson, M.} \emph{et~al.}
\newblock \bibinfo{journal}{\bibinfo{title}{{G}oogle{'}s multilingual neural
  machine translation system: Enabling zero-shot translation}}.
\newblock {\emph{\JournalTitle{Transactions of the Association for
  Computational Linguistics}}} \textbf{\bibinfo{volume}{5}},
  \bibinfo{pages}{339--351}, \doiprefix\url{10.1162/tacl_a_00065}
  (\bibinfo{year}{2017}).

\bibitem{ive-etal-2019-deep}
\bibinfo{author}{Ive, J.}, \bibinfo{author}{Madhyastha, P.} \&
  \bibinfo{author}{Specia, L.}
\newblock \bibinfo{title}{Deep copycat networks for text-to-text generation}.
\newblock In \emph{\bibinfo{booktitle}{Proceedings of the 2019 Conference on
  Empirical Methods in Natural Language Processing and the 9th International
  Joint Conference on Natural Language Processing (EMNLP-IJCNLP)}},
  \bibinfo{pages}{3227--3236}, \doiprefix\url{10.18653/v1/D19-1318}
  (\bibinfo{publisher}{Association for Computational Linguistics},
  \bibinfo{address}{Hong Kong, China}, \bibinfo{year}{2019}).

\bibitem{NEURIPS2019_bdbca288}
\bibinfo{author}{Paszke, A.} \emph{et~al.}
\newblock \bibinfo{title}{Pytorch: An imperative style, high-performance deep
  learning library}.
\newblock In \bibinfo{editor}{Wallach, H.} \emph{et~al.} (eds.)
  \emph{\bibinfo{booktitle}{Advances in Neural Information Processing
  Systems}}, vol.~\bibinfo{volume}{32}, \bibinfo{pages}{8026--8037}
  (\bibinfo{publisher}{Curran Associates, Inc.}, \bibinfo{year}{2019}).

\bibitem{wolf-etal-2020-transformers}
\bibinfo{author}{Wolf, T.} \emph{et~al.}
\newblock \bibinfo{title}{Transformers: State-of-the-art natural language
  processing}.
\newblock In \emph{\bibinfo{booktitle}{Proceedings of the 2020 Conference on
  Empirical Methods in Natural Language Processing: System Demonstrations}},
  \bibinfo{pages}{38--45} (\bibinfo{publisher}{Association for Computational
  Linguistics}, \bibinfo{address}{Online}, \bibinfo{year}{2020}).

\bibitem{adafactor}
\bibinfo{author}{Shazeer, N.} \& \bibinfo{author}{Stern, M.}
\newblock \bibinfo{journal}{\bibinfo{title}{Adafactor: Adaptive learning rates
  with sublinear memory cost}}.
\newblock {\emph{\JournalTitle{CoRR}}}
  \textbf{\bibinfo{volume}{abs/1804.04235}} (\bibinfo{year}{2018}).

\end{thebibliography}

\end{document}